\documentclass{article}

\PassOptionsToPackage{numbers, compress}{natbib}

\usepackage[final]{timeseries_workshop}

\usepackage[utf8]{inputenc} 
\usepackage[T1]{fontenc}    
\usepackage{hyperref}
\usepackage{url}            
\usepackage{booktabs}       
\usepackage{amsfonts}       
\usepackage{nicefrac}       
\usepackage{microtype}      
\usepackage{xcolor}         
\usepackage{enumitem}
\usepackage{graphicx}
\usepackage{tabularx}
\usepackage{amsmath}
\usepackage{amssymb}
\usepackage{booktabs}
\usepackage{wrapfig}
\usepackage{float}
\usepackage{tikz}
\usetikzlibrary{backgrounds}
\usetikzlibrary{arrows,shapes}
\usetikzlibrary{tikzmark}
\usetikzlibrary{calc}
\usepackage{graphicx}
\usepackage{xspace}
\usepackage{tcolorbox}
\usepackage{multirow}

\title{Scaling-laws for Large Time-series Models}

\author{%
  Thomas D. P. Edwards \\
    Johns Hopkins University \\
  \texttt{tedwar42@jhu.edu} \\
  \And
  James Alvey \\
  University of Cambridge \\
  University of Amsterdam \\
  \texttt{j.b.g.alvey@uva.nl} \\
  \AND
  Justin Alsing\\
  Stockholm University \\
  Calda AI \\
  \texttt{justin@calda.ai} \\
  \And
  Nam H. Nguyen \\
  Capital One \\
  \texttt{nam.nguyen@capitalone.com} \\
  \And
  Benjamin D. Wandelt \\
  Institut d’Astrophysique de Paris\\
  CCA, Flatiron Institute \\
  \texttt{bwandelt@iap.fr} \\
}

\begin{document}

\maketitle

\begin{abstract}
Scaling laws for large language models (LLMs) have provided useful guidance in training ever larger models for predictable performance gains. Time series forecasting shares a similar sequential structure to language, and is amenable to large-scale transformer architectures. Here we show that foundational decoder-only time series transformer models exhibit analogous scaling-behavior to LLMs, with architectural details (aspect ratio and number of heads) having a minimal effect over broad ranges. We assemble a large corpus of heterogenous time series data on which to train, and establish for the first time power-law scaling with parameter count, dataset size, and training compute, spanning five orders of magnitude. 
\end{abstract}

\section{Introduction}
\label{sec:intro}
Time-series forecasting is fundamental to decision-making and scientific inference across all domains involving time-ordered observations. In fact, making probabilistic forecasts given past data (whether explicitly or implicitly) arguably underpins every human decision \cite{kording2004bayesian, doya2007bayesian, doya2008modulators, funamizu2016neural, lindig2022bayes}. In industrial and scientific settings, time-series forecasting has traditionally involved supervised training of either statistical models (e.g., ARIMA, GARCH, state-space models, and others; see \citep{west2013bayesian,hyndman2018forecasting} for reviews), bespoke dynamical models based on domain-specific knowledge, or more recently deep-learning based approaches trained for a specific forecasting task (see \citep{torres2021deep} for a review). While these approaches have formed the bedrock of time-series analysis up until now, key challenges and limitations remain: statistical models often fail to describe and capture the latent processes underlying the data, hampering their predictive utility; developing specialized problem-specific models requires considerable investment in human time and resources; and supervised deep-learning approaches trained on a single dataset are typically only useful in the data-rich regime, and generalize poorly to other problems.

The emergence of large language (LLMs; \citep{devlin2018bert,brown2020language,touvron2023llama,chung2024scaling}) and computer vision models \cite{dosovitskiy2020image, radford2021learning, ramesh2021zero, yan2021videogpt, arnab2021vivit, he2022masked, li2023blip} with zero-shot prediction capabilities has sparked an interest in developing foundation models for time-series — general purpose forecasting models, pre-trained on a large and diverse corpus of time-series data, aimed at achieving state-of-the-art (SOTA) zero-shot forecasting performance across many domain areas~\cite{2023arXiv231010688D, 2024arXiv240203885G,2023arXiv231008278R, 2023arXiv231003589G, 2022arXiv221114730N, 2024arXiv240202592W, 2023arXiv231005063W, 2023arXiv230512095X, 2024arXiv240210198I, salinas2020deepar, oreshkin2019n, oreshkin2021meta, gruver2024large, ma2023survey, 2024arXiv240307815F,  2023arXiv230514406K}. Large time-series models (LTMs) are already achieving zero-shot prediction capability similar to or better than baseline statistical or domain-specific models in many areas \cite{2023arXiv231010688D, 2024arXiv240203885G,2023arXiv231008278R, 2023arXiv231003589G, 2022arXiv221114730N, 2024arXiv240202592W, 2023arXiv231005063W, 2023arXiv230512095X, 2024arXiv240210198I, 2024arXiv240307815F}.

Underpinning the investment and subsequent success of LLMs and large-scale computer vision models was the demonstration of neural scaling laws \citep{originalscaling, tan2019efficientnet, tan2021efficientnetv2, raghu2021vision, riquelme2021scaling, he2022masked, henighan2020scaling}. The observed power-law scaling of test loss with model size, compute resources, and training set size, has provided a basis for predicting expected gains from different efforts, aiding the community in allocating resources appropriately to achieve performance breakthroughs. Given qualitative differences in data and modeling challenges, existence of neural scaling-laws for time-series is not guaranteed from the language and computer vision results. Establishment of similarly favourable scaling laws for LTMs would serve as a motivation and guide in the pursuit of foundation models for time-series forecasting.

\textbf{Contributions:} We establish for the first time that LTMs enjoy similar power-law scaling laws to language and computer vision. We train decoder-only transformer models (with architectures tailored to time-series forecasting; \S\ref{sec:data_methods}) on a large, diverse, and well-balanced dataset comprising around 8 billion data points across 30,211,687 individual time-series, drawn from $38$ qualitatively distinct data sources from varied areas (see \S \ref{sec:data_methods}). We demonstrate power-law like scaling behavior of model performance with model size, compute, and dataset size (Fig. \ref{fig:scaling}), finding similar scaling behavior in three key measures of model performance: the mean-square error (MSE) characterizing the accuracy of point (posterior mean) forecasts; the Continuous Ranked Probability Score (CRPS \cite{hersbach2000decomposition}) characterizing the fidelity of the probabilistic predictions (ie., coverage of the forecast posterior density); and the log-likelihood loss characterizing the Kullback-Leibler (KL) divergence between the model and data generative distributions.
\section{Data and Methods}
\label{sec:data_methods}
\textbf{Data:} The development of a foundation model for time-series forecasting is predicated on the availability of a sufficiently large, diverse, and well-balanced dataset to train on. We constructed a corpus of time-series data comprising around $8$ billion data points drawn from $38$ varied data sources (see Tab.~\ref{tab:dataset_stats}). For the purpose of this study, our focus was to ensure our dataset is: large enough so that for our largest models ($\sim 100$M parameters) we are still operating in the $\sim$infinite data limit (c.f. \cite{originalscaling}); as diverse as possible given the practical limitations on publicly available data; and well-balanced, such that no individual dataset comprises more than roughly $15\%$ of the total number of data points. Our resulting dataset is competitive with the SOTA in terms of both diversity and size\footnote{Where other SOTA time-series datasets from recent studies are larger, this discrepency is mostly accounted for by the use of synthetic data (which we deliberately do not include), or reduction in our total data count from re-balancing the data to ensure it is not dominated by a single source.}, while covering a wide variety of sampling frequencies, record lengths, dynamic ranges, and underlying latent process phenomenology. We focus exclusively on univariate time-series forecasting, and leave the study of scaling-laws for multivariate LTMs to future work. Data sources, balancing procedure and pre-processing steps are detailed in Appendix \ref{app:data}.

\textbf{Model Architecture:} We use decoder-only transformer models with self-attention as the primary architecture throughout, with a context length of $256$ data points and ReLU activation functions. Following the performance gains shown in Ref.~\cite{2023arXiv231005063W}, we use a learnable encoding rather than the sinusoidal positional encoding used in the original transformer model \cite{vaswani2017attention}. Both the learned positional encoding and embedding are simple linear layers going from one input to $d_\mathrm{m}$ outputs.

\textbf{Distribution Head:} We use a Student's-$t$ distribution head, where the mean, variance, and degrees of freedom are each modelled by a separate dense network with four hidden layers of dimension $d_\mathrm{m}$. The Student's-$t$ distribution allows us to model heavy-tailed data, which we find in experiments enables significantly more stable training than a Gaussian head or MSE loss. We note that in reality, time-series data and processes exhibit diverse distributional characteristics, and a more expressive distribution head (e.g., mixture model, normalizing flow or diffusion model) is well-motivated. We leave the exploration of scaling-laws under more expressive distribution heads to future work. We use a negative log-likelihood (KL) loss for training throughout. 

\textbf{Parameter Counting:} With this setup, the model architecture is defined by the following parameters: the number of output dimensions $\theta_{\mathrm{out}}$, the input/output size of the linear layers in the self-attention $d_\mathrm{m}$, the number of heads $N_{\mathrm{heads}}$, the hidden layer size of the linear layers directly after the self-attention $d_{\mathrm{ff}}$, and the number of decoder layers $N_\mathrm{l}$. Throughout this work we fix $d_\mathrm{m} = d_{\mathrm{ff}}$ and treat all trainable parameters (including weights and biases of all layers) equally in the parameter counting. As shown in Fig.~\ref{fig:scaling}, we explore models with $\sim 10^3$ to $\sim 10^8$ trainable parameters.

\textbf{Learning rate and architecture sensitivity:} To extract reliable scaling laws, we need to determine sensitivity and robustness to the learning rate (LR) schedule and architecture choices. We use a linear warm up followed by sinusoidal decay for the learning rate scheduling, and find that the model performance clearly depends on the maximum LR reached at the end of the warm-up. To ensure robustness to the maximum LR, we fit a power-law to the best model at each parameter size to estimate the optimal maximum LR as a function of parameter count, shown in Fig. \ref{fig:LR_scaling} (Appendix \ref{app:lr_dep}).

Figure~\ref{fig:model_shape_scaling} (Appendix \ref{app:lr_dep}) shows how the minimum CRPS varies as a function of aspect ratio $d_m/N_\mathrm{l}$ (left panel) and the number of attention heads, $N_{\mathrm{heads}}$ (right panel). Performance is $\sim$insensitive to the number of heads, and only weakly sensitive to aspect ratio for aspect ratios $\lesssim100$ (after which performance drops steeply). We note that this is analogous to the weak architecture sensitivity observed for LLMs \cite{originalscaling}. For the main parameter-, compute-, and data-scaling runs, we fix the number of heads to four, and keep the aspect ratio $<70$. See Appendix \ref{app:training} for further training details.
\section{Results}
\label{sec:results}
\begin{figure}[t!]
    \centering
\includegraphics[width=0.96\linewidth,trim={0.1cm 0.1cm 0.1cm 0.1cm},clip]{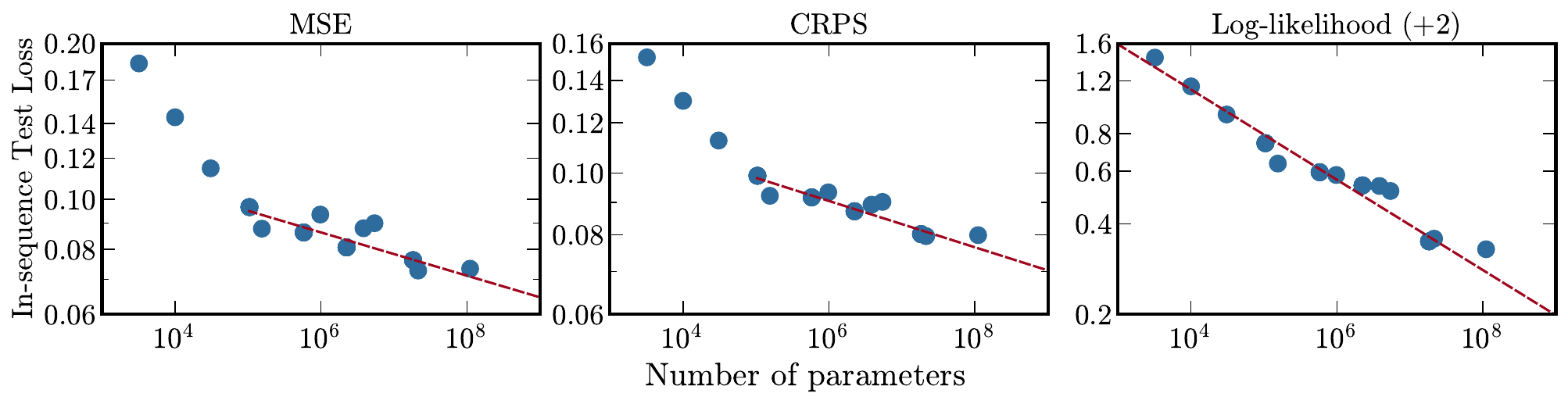}
\includegraphics[width=0.96\linewidth,trim={0.1cm 0.1cm 0.1cm 0.1cm},clip]{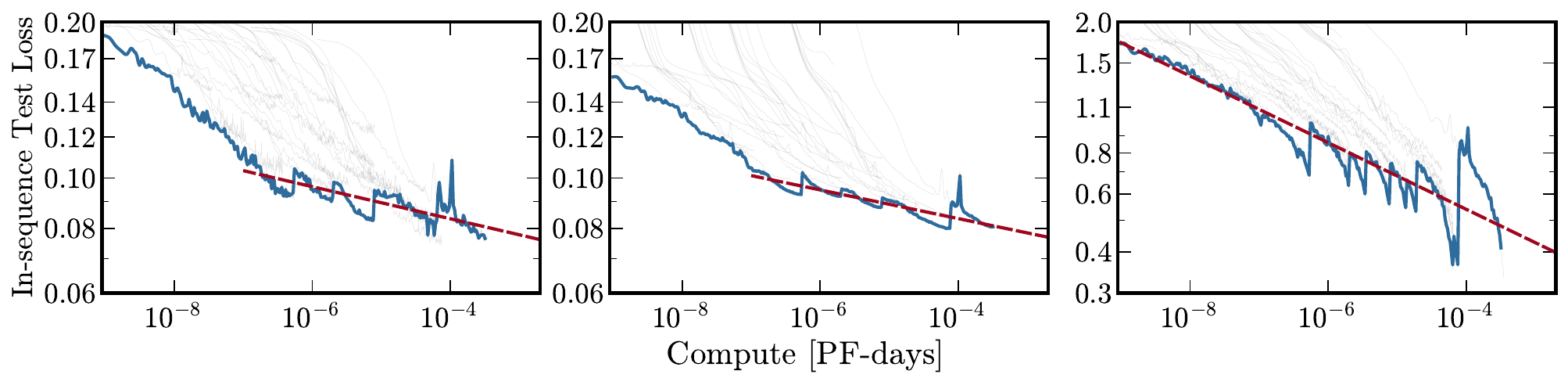}
\includegraphics[width=0.96\linewidth,trim={0.1cm 0.1cm 0.1cm 0.1cm},clip]{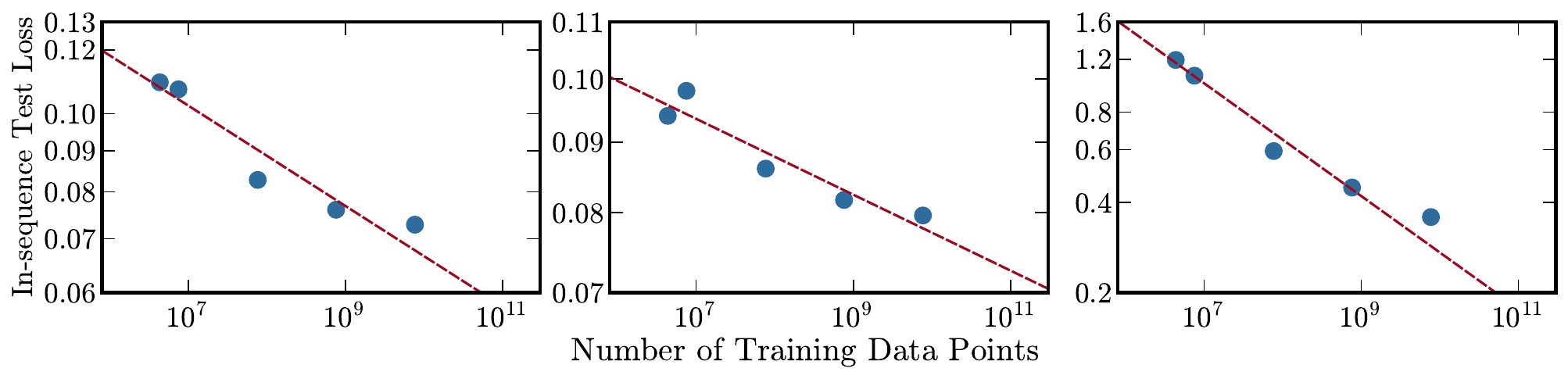}
    \caption{\textbf{Test Loss Scaling Laws:} Minimum MSE (left), CRPS (middle) and log-likelihood (right)  in-sequence test metrics as a function of the number of parameters (top), compute (middle), and dataset size (bottom).}
    \label{fig:scaling}
\end{figure}
Scaling as a function of parameter count $N_p$, dataset size $\mathcal{D}$, and compute $\mathcal{C}$ is summarized in Fig.\ref{fig:scaling}. For each scaling-relation, we fit a power law of the form $\ln L(A) = -B_0\ln A + B_0 \ln A_0$,
where $L$ is the objective function (MSE, CRPS, or log-likelihood) and $A$ is the scaled quantity (i.e., parameter count, dataset size, or compute). The fitted parameter values are given in Tab.~\ref{tab:fits} (Appendix \ref{app:pl_fits}). Where broken power-law like scaling is observed, we report the power law fit after the break only, since this is the relevant quantity to motivate extrapolation to larger models / datasets / compute resources. 

\textbf{Parameter scaling:} Fig.~\ref{fig:scaling} (top row) shows the minimum in-sequence test loss (MSE, CRPS, and log-likelihood\footnote{
    We add a constant factor of two to the log-likelihood to ensure values are always positive, enabling us to examine power-law scaling. A constant additive factor can change the slope of the fitted power-law; to remain as agnostic as possible we choose to add the smallest integer required to make all values of the loss positive.}) as a function of parameter count, showing $\sim$power-law behavior over nearly five orders of magnitude in model size. A mild break is observed in the power-law behavior in both the MSE and CRPS test losses, indicating qualitatively different behavior for smaller models. In contrast, little or no break is seen in the log-likelihood scaling; this qualitative difference relative to MSE and CRPS is likely due to the log-likelihood being more sensitive to variations in the tails of the forecast distribution (see e.g., ~\cite{BJERREGARD2021100058}). The observed scaling over many orders of magnitude demonstrates that LTMs are likely to reach SOTA performance given enough data and model size.

\textbf{Data Scaling:} Extracting reliable scaling behavior with dataset size requires keeping the data-diversity fixed, i.e., each dataset's relative contribution to the total data count should remain the same under scaling (see Tab.~\ref{tab:dataset_stats}). For time-series that are significantly longer than our context length, we use a randomly chosen portion, $f_d$, of each time-series, while for series that would become shorter than our context length once cut, we instead randomly drop the entire series with probability equal to $1-f_d$. We compute the test loss over the full test set to allowing direct comparison between runs with different values of $f_d$, and reduce the noise on the test loss in the small (scaled) dataset limit. 

Results are shown in the bottom row of Fig.~\ref{fig:scaling} where we train a $\sim 20$M parameter model using the optimum max learning rate found during the parameter scaling exploration, with early stopping. We find power-law scaling across four orders of magnitude in all three performance measures.

\textbf{Compute Scaling:} The compute at any given stage in the training process is given by $\mathcal{C} = 6BN_\mathrm{p}L_\mathrm{seq}$, where $B$ is the batch size, $N_\mathrm{p}$ is the number of parameters in the model, and $L_\mathrm{seq}$ is the context length \cite{originalscaling}. Test losses as a function of compute are shown in Fig.~\ref{fig:scaling} (middle row), where the scaling law is obtained from the minimum test loss attained at any given value of $\mathcal{C}$. Although we see a significant amount of noise in the loss functions during training, there is a clear overall trend towards lower test losses for higher compute, which is well-described by a power law. Similarly to the parameter scaling we see a mild break at low compute values for both the MSE and CRPS test losses. Note that while the MSE and CRPS metrics appear to be approximately converged over the compute range considered, the log-likelihood may not be fully converged; additional training may be needed to obtain accurate compute scaling-law fits for the log-likelihood in the large model limit. 
\section{Discussion}
\label{sec:discussion}
We have focused on evaluating models by their in-sequence (next step) test loss, rather than explicitly assessing model's ability to forecast further into the future. The implicit assumption that a model with good in-sequence predictions should naturally be able to forecast into the future is theoretically and empirically well-motivated: as the modelled posterior predictive distribution for the next value improves, any accumulated errors from auto-regressive roll-out those predictions into the future should also improve. In App.~\ref{app:forecast}, we show some clear examples of how forecast roll-out becomes increasingly coherent with increasing model size. We leave the study of scaling-laws based on forecasting ability on different time horizons to future work.

We have detailed the specific scaling laws for a decoder-only transformer with self-attention. However, it would be interesting to explore how modifications to this architecture might improve model scaling. In particular, much of the recent progress in using LTMs~\cite{2023arXiv231010688D, 2024arXiv240203885G,2023arXiv231008278R, 2023arXiv231003589G, 2022arXiv221114730N, 2024arXiv240202592W, 2023arXiv231005063W, 2023arXiv230512095X, 2024arXiv240210198I, 2024arXiv240307815F} has involved various changes to transformer architectures to make them more suited to time-series data. We advocate for comparative scaling law studies as new architectures are introduced, to allow the community to evaluate which model architectures will eventually reach SOTA zero-shot prediction capabilities.

When experimenting with data scaling, we found it was critical to scale the training data in such a way as to preserve the data diversity; approaches to data scaling that did not preserve data diversity failed to reveal any clear scaling behaviour. Given the importance of data diversity in establishing data scaling laws, and in training SOTA pre-trained foundation models in general, developing a robust framework for quantifying data diversity would be of great utility to the field.

One scaling law that we have not explored in this work (due to computational limitations) is performance as a function of increasing context length. Multiple studies (e.g., \cite{2024arXiv240307815F, 2024arXiv240305530G}), both for LLMs and LTMs, have shown that increasing the context length significantly improves both in-sequence prediction and forecasting, and a recent study \cite{2024arXiv240515124S} find interesting scaling behaviour of LTMs with context length. We will explore context-length scaling in future work.

We have focused on univariate time-series data. However, a general purpose foundation model for time-series forecasting should be able to cope with the more general setting of multivariate time-series prediction, with multiple exogeneous covariates. Establishing scaling-laws for multivariate time-series forecasting will be an important extension to this work; this demands the assembly of a large and diverse training set of multivariate data, each with their own exogeneous factors.



\bibliographystyle{ieeetr}
\bibliography{main.bib}


\appendix

\newpage
\section{Dataset Details}
\label{app:data}

\begin{table}[t!]
\renewcommand{\arraystretch}{1.2}
\centering
\caption{\textbf{Dataset summary}. M indicates million and B indicates billion.}
\label{tab:dataset_stats}
\begin{tabularx}{\textwidth}{l*{7}{>{\centering\arraybackslash}X}}
\toprule
& \textbf{Monash} & \textbf{Climate} & \textbf{Energy} & \textbf{Traffic} & \textbf{Finance} & \textbf{Audio} & \textbf{Total} \\ \midrule
\textbf{Datasets} & 23 & 15 & 2 & 5 & 2 & 3 & 38 \\
\textbf{\# of data points} & 503M & 1.56B & 2.5B & 1.5B  & 42.6M & 1.98B & 8.13B \\
\textbf{\% of data} & 6.18\% & 19.19\% & 30.75\% & 18.45\% & 0.52\% & 24.35\% 
& 100\% \\ \bottomrule
\end{tabularx}
\end{table}

In this section we detail the various sources that form the basis of our dataset and the choices made during its construction and re-balancing. Constructing a training set for establishing foundational scaling relations requires three key considerations. Firstly, the dataset should be large enough so that for the largest models trained, we are still operating in the $\sim$infinite data limit (see e.g., \citealp{originalscaling}). Secondly, the dataset needs to be sufficiently diverse so that any results are representative of the foundation-model regime, covering a large volume of the space of time-series phenomenology. Thirdly, the dataset needs to be balanced, so that any scaling results are representative of foundation model behaviour and not tied to performance gains for a single or handful of dominant dataset(s).

Taking inspiration from large language models~\cite{originalscaling}, we therefore aimed to gather around $\mathcal{O}(10^{10})$ data points from a variety of domains. We note that treating a single floating point number on a similar footing to a language token is not necessarily a good comparison; language tokens can contain significantly more semantic meaning than a floating point number can. The continual growth of open-source time-series datasets in both size and diversity will enable increasingly robust neural scaling studies.

Before detailing our particular sources, we would like to emphasize that there is a large corpus of time series data publicly available but is not currently formatted for easy downloading and processing. Ref.~\cite{2024arXiv240202592W} was the first paper to open-source a large dataset\footnote{
    Note that by the time this data became open-source we had already fixed our dataset for the production runs completed for this study.
}, setting a trend for improved training and benchmarking of foundational time-series models. However, significant work is still required to expand available datasets in size and diversity to reach the same maturity as LLMs (large-scale, SOTA language models are trained on well over a trillion tokens). 

We now discuss each dataset presented in Tab.~\ref{tab:dataset_stats}.

All data used throughout this work has been labelled/licensed as free to use for non-commercial purposes with the appropriate citations. We have included the appropriate citations where necessary below.

\subsection{Monash}

The Monash dataset has been the default repository of open-source time series data used by the academic community for some time~\cite{godahewa2021monash}. It contains data from a huge variety of sources and contains a wide variety of characteristics. For this work we exclude series that are either too short or are particularly noisy.\footnote{
    We found through experimentation that removing very noisy datasets significantly improved training stability.
}
We are then left with a total of 23 different sources which add up to a total of $\sim 500\mathrm{M}$ data points; details are given in Tab.~\ref{tab:monash}.

\begin{table}[h!]
\renewcommand{\arraystretch}{1.2}
\centering
\caption{\textbf{Monash Data:} For each dataset we list the sampling frequency, the total number of series, and the total number of data points.}
\label{tab:monash}
\begin{tabularx}{\textwidth}{l|*{3}{>{\centering\arraybackslash}X}}
\toprule
\textbf{Dataset} & \textbf{Frequency} & \textbf{Number of Series} & \textbf{Number of Data Points} \\ \midrule
London Smart Meters & Half Hourly & 5,560 & 166.5M \\
Wind Farms & Every Minute & 339 & 172.1M \\
Wind Power & 4 Seconds Intervals& 1 & 7.4M \\
Solar Power & 4 Second Intervals & 1 & 7.4M \\
Oikolab Weather & Hourly & 8 & 0.8M \\
Elecdemand & Half Hourly & 1 & 17.5k \\
Kaggle Web Traffic & Daily & 145,063  & 116.5M \\
Tourism Quarterly & Quarterly & 427 & 42.5k \\
Tourism Monthly & Monthly & 366 & 109.3k \\
CIF 2016 & Monthly & 72 & 7.1k \\
Traffic Weekly & Weekly & 862 & 89.6k \\
Traffic Hourly & Hourly & 862 & 15.1M \\
Australian Electricity & Half Hourly & 5 & 1.2M \\
Sunspot & Daily & 1 & 73.9k \\
Hospital & Monthly & 767 & 64.4k \\
NN5 Daily & Daily & 111 & 87.8k \\
NN5 Weekly & Weekly & 111 & 12.5k \\
M4 Hourly & Hourly & 414 & 373.4k \\
Fred MD & Monthly & 107 & 77.9k \\
Solar Weekly & Weekly & 137 & 7.1k \\
Solar 10 Minutes & 10 Minute Intervals & 137 & 7.2M \\
Electricity Weekly & Weekly & 321 & 50.1k \\
Electricity Hourly & Hourly & 321 & 8.4M \\
\bottomrule

\end{tabularx}
\end{table}

\subsection{Climate}

Our climate dataset, made up of around 1.5B data points, has two primary sources: the National Oceanic and Atmospheric Administration (NOAA) and the fifth generation European Centre for Medium-Range Weather Forecasts atmospheric reanalysis of the global climate (ERA5). Each source provides approximately 750M data points split across a variety of observables and time frames.

We note here that since the global climate is a correlated system, forecasting a single variable into the future whilst ignoring the evolution of the rest of the system is intrinsically difficult (maybe impossible in some cases). Nevertheless, each time series can provide important information from which the foundation model can learn correlations. Moreover, some seasonal trends are very stable and predictable from a single time series. Future work should carefully consider how to include climate data in a way that allows the model to exploit correlations inherent in the data~\cite{graphcast, fourcastnet}.

\begin{table}[h!]
\renewcommand{\arraystretch}{1.2}
\centering
\caption{\textbf{NOAA Data:} For each dataset we list the sampling frequency, the total number of series, the length of each series, and the total number of data points.}
\label{tab:NOAA}
\begin{tabularx}{\textwidth}{l|*{4}{>{\centering\arraybackslash}X}}
\toprule
\textbf{Dataset} & \textbf{Frequency} & \textbf{Number of Series} & \textbf{Length} & \textbf{Number of Data Points} \\ \midrule
SST Mean & Daily & 582241 & 365 & 212.5M  \\
SST Anomalies & Daily & 581249 & 365 & 212.1M  \\
SST Long Term Average & Daily & 218211 & 365 & 79.6M  \\
SST Monthly Average & Monthly & 72730 & 509 & 37M  \\
SST Weekly Average & Monthly & 72689 & 2214 & 161M  \\ \midrule

Ice Mean & Daily & 63971 & 365 & 23M  \\
Ice Long Term Average & Daily & 12451 & 365 & 4.5M  \\
Ice Monthly Average & Daily & 5363 & 509 & 2.7M  \\ \midrule

Radiation Long Term Average & Daily & 6622 & 365 & 2.4M  \\ \bottomrule

\end{tabularx}
\end{table}

\vskip 6pt

\textbf{NOAA:} We primarily gather data from NOAA high-resolution blended analysis of daily sea surface temperature (SST) which includes both temperature measurements and ice level measurements on a $0.25^{\circ}$ grid worldwide.\footnote{
    The original data can be found here \url{https://downloads.psl.noaa.gov/Datasets/noaa.oisst.v2.highres/}.
} Weather at different points of the grid are intrinsically correlated, especially on such small grid sizes. We therefore downsample the data by a factor of three by randomly choosing grid points without replacement (we do this independently for each dataset). 

To ensure we have data that covers a wide range of time scales and variability we pick a variety of observables shown in Tab.~\ref{tab:NOAA}. For the daily data we pick 8 years of data, each separated by 5 years (spread out to maximize data diversity i.e., minimize year to year correlations) but skip leap years for easier data processing (so all arrays are 365 elements long). The final year selection is 1985, 1990, 1995, 2001, 2005, 2010, 2015, and 2021. This size of this dataset could easily be supplemented simply by adding more of the 40 years of available data.

For additional diversity we use the same method to extract outgoing long wave radiation time series from \url{https://downloads.psl.noaa.gov/Datasets/uninterp_OLR/}. This is the shown in the final row of Tab.~\ref{tab:NOAA}.

\begin{table}[h!]
\renewcommand{\arraystretch}{1.2}
\centering
\caption{\textbf{ERA5 Data:} Similar to above. The different number of series for each dataset is due to the randomness in the subsampling.}
\label{tab:ERA5}
\begin{tabularx}{\textwidth}{l|*{4}{>{\centering\arraybackslash}X}}
\toprule
\textbf{Dataset} & \textbf{Frequency} & \textbf{Number of Series} & \textbf{Length} & \textbf{Number of Data Points} \\ \midrule
Sea Level Pressure & 4 Hour Intervals & 63094 & 2190 & 138M \\
2m Temp. & 4 Hour Intervals & 63190 & 2190 & 138M \\
2m Dewpoint Temp. & 4 Hour Intervals & 63123 & 2190 & 138M \\
Surface Pressure & 4 Hour Intervals & 63263 & 2190 & 139M \\
10m V Wind Comp. & 4 Hour Intervals & 63263 & 2190 & 139M \\ 
10m U Wind Comp. & 4 Hour Intervals & 63220 & 2190 & 138M \\ \bottomrule

\end{tabularx}
\end{table}
\textbf{ERA5:} We take a similar approach to above when processing/gathering ERA5 data. Here though, we focus on higher frequencies by using a single year of data (2001) sampled every four hours. We additionally use different data variables (the six most popular variables) to ensure that the data features are likely different to those present in the NOAA data. ERA5 data is also originally on a $0.25^{\circ}$ global grid which we randomly down sample by a factor of four. Details are given in Tab.~\ref{tab:ERA5}.

\subsection{Energy}

For the energy dataset, we use the benchmark dataset prepared in the \texttt{BuildingsBench} data release~\cite{emami2023buildingsbench}. In particular, we choose to sample 2.5B data points from the full dataset (which totals over 15B individual data points). These 2.5B data points, which overall constitute approximately 30\% of our full dataset, are all taken from the Buildings-900K database. These time series represent a large-scale sample of simulated US building energy demand and are designed to be broadly representative of US commercial and residential building stock. As described in~\cite{emami2023buildingsbench}, the dataset is originally sourced from the NREL EULP database~\cite{osti_1854582}, which provides 15- minute resolution, appliance-level consumption for 550K residential and 350K commercial buildings spread across all climate regions in the U.S. For more finer-grained details, see App.~B.3 in Ref.~\cite{emami2023buildingsbench}.

\subsection{Traffic}

We consider the public LargeST~\cite{liu2023largest} dataset which is a collection of 8600 time series recorded from traffic sensors in the California area. The data spans over 5 years, from 2017 to 2021, and is sampled at 15 minute resolution. To reduce the data size, we down-sample the data to hourly resolution and remove series that contains over $50\%$ missing entries. This gives us a total of 8520 series all with length 175296, which translates to 1.46B data points.

\subsection{Finance}

We include daily stock returns and volume data, treated as separate one-dimensional time-series respectively, for $5038$ stocks listed across the Nasdaq, NYSE, and AMEX stock exchanges. Daily stock returns and volume tickers are obtained for $7230$ stocks from \texttt{yahoo finance}, from the beginning of each listing up to 1st January 2024. We discard any stocks that have fewer than $512$ ticks (recorded trading days), and any series containing \texttt{NaN} or \texttt{inf}. This results in time-series for $5038$ stocks, with both returns and volume data, and a total of $42.6$M data points (Tab. \ref{tab:finance}).
\begin{table}[h!]
\renewcommand{\arraystretch}{1.2}
\centering
\caption{\textbf{Finance Data:} Daily stock returns and volume data for $5038$ stocks listed across the Nasdaq, NYSE and AMEX exchanges, obtained from \texttt{yahoo finance}.}
\label{tab:finance}
\begin{tabularx}{\textwidth}{l|*{3}{>{\centering\arraybackslash}X}}
\toprule
\textbf{Dataset} & \textbf{Frequency} & \textbf{Number of Series} & \textbf{Number of Data Points} \\ \midrule
Stock Returns & Daily & 5038 & 26.3M  \\
Stock Volume & Daily & 5038 & 26.3M  \\
\bottomrule
\end{tabularx}
\end{table}

\subsection{Audio}

Audio data is intrinsically a one dimensional time series rich with structure and features; it is therefore perfectly suited for our study.  We have three primary sources of audio data, all from the DagsHub Open-Source Audio Datasets repository (\url{https://github.com/DagsHub/audio-datasets}). Again, the total volume of data here is extremely large and can be used to supplement future datasets for larger models. Here we use three particular sources each from a different domain to enhance its diversity. As presented in Tab~\ref{tab:dataset_stats}, these three sources add up to approximately 2B data points and $\sim 25\%$ of our overall dataset. A summary of the three sources can be found in Tab.~\ref{tab:audio}

\begin{table}[h!]
\renewcommand{\arraystretch}{1.2}
\centering
\caption{\textbf{Audio Data:} Similar to above.}
\label{tab:audio}
\begin{tabularx}{\textwidth}{l|*{4}{>{\centering\arraybackslash}X}}
\toprule
\textbf{Dataset} & \textbf{Frequency} & \textbf{Number of Series} & \textbf{Length} & \textbf{Number of Data Points} \\ \midrule
Commands & 16 $\mathrm{kHz}$ & 47650 & 16,000 & 762.4M  \\

Arabic Speech & 24 $\mathrm{kHz}$ & 1813 & Varied & 329.9M  \\

Bird Audio & 22 $\mathrm{kHz}$  & 4000 & Varied & 888.3M  \\ \bottomrule

\end{tabularx}
\end{table}

\textbf{Commands:} The speech command dataset~\cite{2018arXiv180403209W} is made up of a series of short audio files with different voices saying a collection of common English words (e.g., ``happy'' and ``five''). From all the data provided \url{https://github.com/DagsHub/audio-datasets/blob/main/Speech_Commands_Dataset/README.md} we take a random half of the data and exclude any clips that are not 16k long (again for easy saving). We are then left with 47650 series, making a total of $\sim 750\mathrm{M}$ data points.

\vskip 3pt

\textbf{Arabic Speech:} This dataset contains 1813 time series of high quality (studio recorded) spoken Arabic utterances sampled at $48\mathrm{kHz}$ -- \url{https://github.com/DagsHub/audio-datasets/tree/main/Arabic-Speech-Corpus}. To reduce the data size without dramatically affecting its quality, we down sample the data by a factor of two (human speech is typically below $24\mathrm{kHz}$). This gives us a total of $\sim 300\mathrm{M}$ data points.

\vskip 3pt

\textbf{Birds:} Finally, we use the bird detection dataset from \url{https://github.com/DagsHub/audio-datasets/blob/main/Bird-Audio-Detection-challenge/README.md}~\cite{https://doi.org/10.1111/2041-210X.13103}. This dataset contains a combination of bird and other sounds designed to train machine learning algorithms to detect bird noises. Here we ignore the labels and use the entire dataset in training. Again, to reduce data volumes we down sample by a factor of two, and only use a randomly chosen half of the data. This leaves us with 4000 time series sampled at 22 $\mathrm{kHz}$ for a total of $\sim 900\mathrm{M}$ data points.

\begin{figure}[t!]
    \centering
\includegraphics[width=\linewidth,trim={0.1cm 0.1cm 0.1cm 0.1cm},clip]{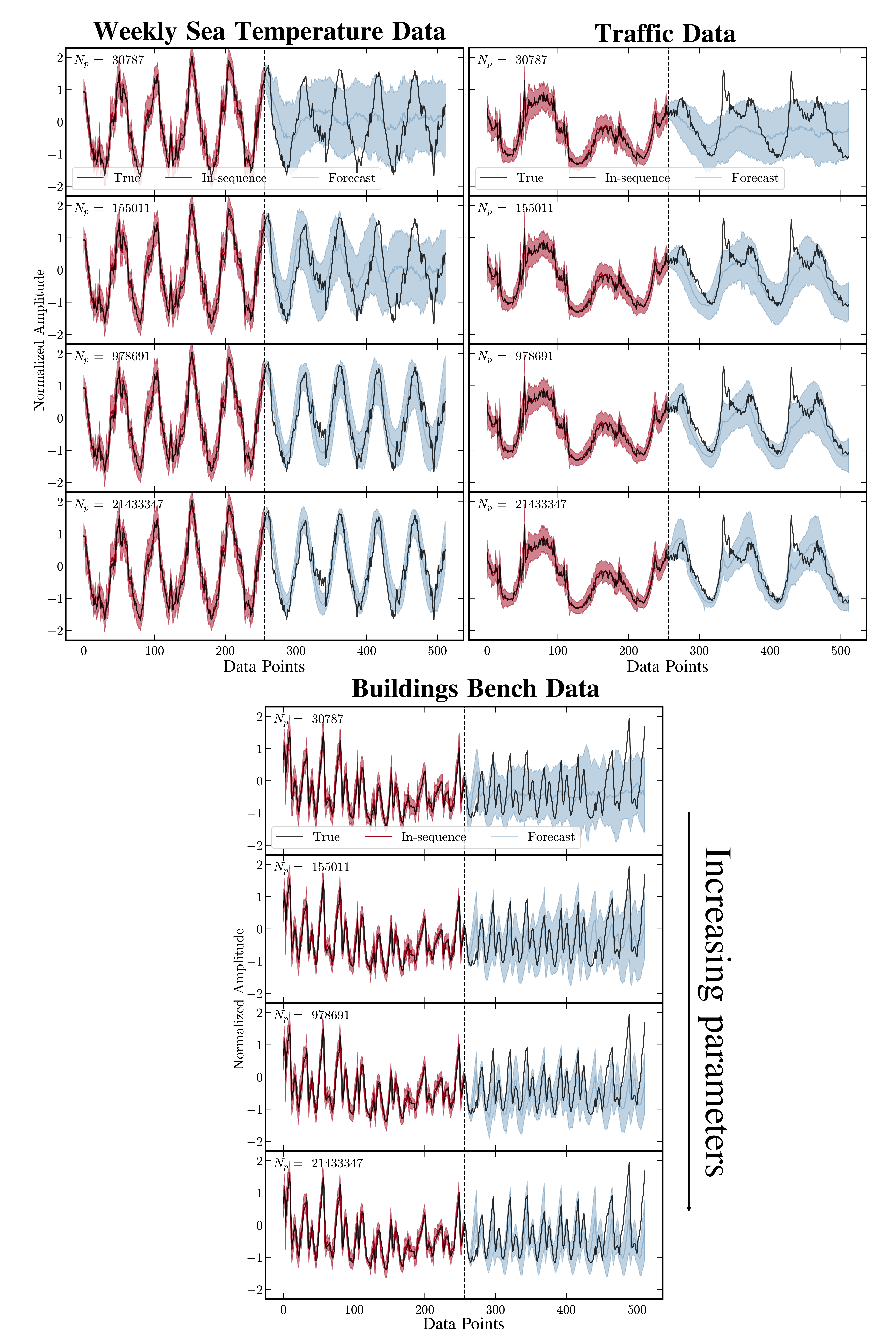}
    \caption{\textbf{In-sequence test loss to forecasting:} Here we show the connection between improved in-sequence test loss and forecasting performance as a function of model size. In particular, we show the true data in black with $1\sigma$ ranges for both in-sequence and forecasting predictions. It is clear that as in-sequence test loss decreases, forecasting also becomes substantially more predictive.}
    \label{fig:forecast}
\end{figure}

\subsection{Dataset balancing and pre-processing}

Each dataset is made up of a large number of individual time series of varying lengths. We use 95\% of the set of time series for training and the remaining 5\% for testing. Since the majority of the series are significantly longer than our context window, during training and testing we visit each series with probability $p_i = t_i/T$, where $t_i$ is the number of data points in that series and $T$ is the total number of data points in the training set. Additionally, each time we visit a series we choose a random starting index. This strategy ensures that the model sees each section of the data once (on average) in a given epoch. We normalize each time-series in the training set to have zero mean and unit standard deviation.\footnote{In rare instances where input time-series are constant (and hence have zero standard deviation), we set them to a constant value of zero.}

\section{Training details and compute requirements}
\label{app:training}
We use the \texttt{AdamW} optimizer with a batch size of 512, a cosine learning rate scheduler with a linear warm up of 3000 training steps, and train for a total of $10^5$ steps. When training on the entire dataset ($\sim8$B data points), this equates to roughly two epochs. To reduce computational costs we compute the test loss every $\mathcal{O}(200)$ steps and average over a random 10\% of the test data each time.

To produce the results in this paper requires $\mathcal{O}(50 - 70)$ individual production runs. Apart from the 100M parameter run, these were all carried out on single A100 NVIDIA GPU instances, each taking between 1 and 3 days to complete. As such, overall, the work presented here required $\mathcal{O}(150)$ GPU-days of compute. To host the full dataset, we also required a CPU RAM allocation of approximately 250 GB.

\section{Learning-rate and architecture dependence}
\label{app:lr_dep}
\begin{figure}[t!]
    \centering
\includegraphics[width=0.7\linewidth,trim={0.1cm 0.1cm 0.1cm 0.1cm},clip]{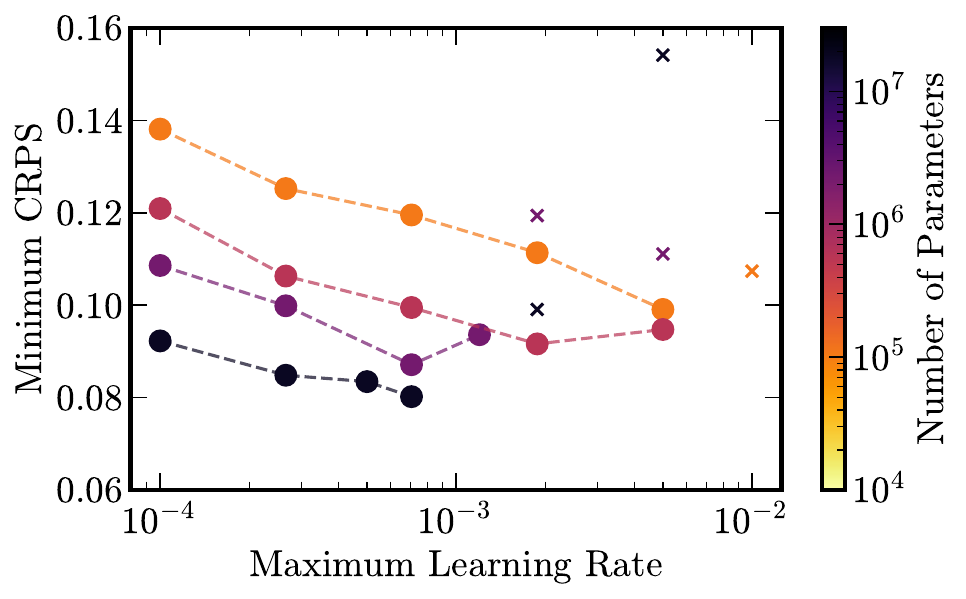}
    \caption{\textbf{Importance of Learning Rate:} Here we show the minimum CRPS measured on the test data as a function of the maximum learning rate reached at the end of the linear warm up schedule. Crosses indicate that the model diverged before training was complete. There is a clear optimum max learning rate which decreases as a function of model size/number of parameters.}
    \label{fig:LR_scaling}
\end{figure}

In Fig.~\ref{fig:LR_scaling} we show the effect of changing the maximum learning rate reached at the end of the warm up. The performance of the model (CRPS) clearly depends on the maximum learning rate, and that dependence is itself a function of parameter count. The dependence on maxmimum learning rate is strong enough that it is possible to get better performance with a smaller model, if the maximum learning rate is too small (or too large) for the larger model. Moreover, for a fixed model size we see a clear optimum learning rate above which models diverge (shown as crosses on       Fig.~\ref{fig:LR_scaling}). To ensure that we used an optimal maximum learning rate as a function of model size, we fit a power law with a constant offset to the best models (at each parameter size) shown in Fig. ~\ref{fig:LR_scaling}. In the few cases for the largest models where our power law fit overestimates the optimal maximum learning rate (leading to divergence), we slowly reduce the learning rate until we achieve convergence.

\begin{figure}[t!]
    \centering
\includegraphics[width=\linewidth,trim={0.1cm 0.1cm 0.1cm 0.1cm},clip]{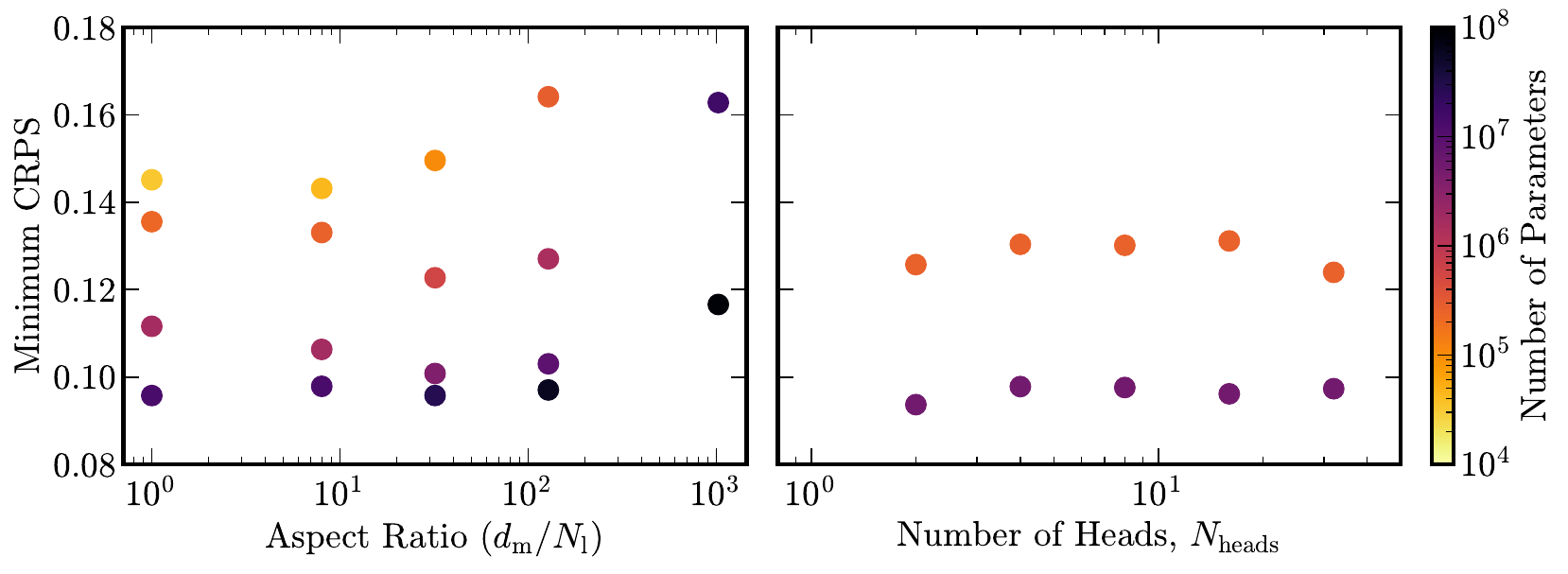}
    \caption{\textbf{Importance of Transformer Architecture:} We show the minimum CRPS on the test set as a function of architecture choices and number of parameters. \textit{Left:} Performance on the test data has a weak dependence on aspect ratio below $<100$ but degrades significantly $>128$. We therefore keep aspect ratios $<70$ for all scaling runs. \textit{Right:} Here we see that the number of attention heads has no noticeable affect on the performance for both model sizes tested. We fix the number of heads to four for the scaling runs.} 
    \label{fig:model_shape_scaling}
\end{figure}

In Figure \ref{fig:model_shape_scaling} we show the dependence of model performance on architecture choices, showing that performance is largely invariant to the number of heads, while being only weakly sensitive to the aspect ratio (for aspect ratios $\lesssim 100$).

\section{Power-law fits}
\label{app:pl_fits}
In \ref{tab:fits} we provide the power law fits to teh scaling relations shown in Fig.~\ref{fig:scaling}.
\begin{table}[h!]
\renewcommand{\arraystretch}{1.2}
\centering
\caption{\textbf{Power-law fits.}}
\label{tab:fits}
\begin{tabularx}{0.9\textwidth}{l|*{2}{>{\centering\arraybackslash}X}|*{2}{>{\centering\arraybackslash}X}|*{2}{>{\centering\arraybackslash}X}}
\toprule
& \multicolumn{2}{c|}{MSE} & \multicolumn{2}{c|}{CRPS} & \multicolumn{2}{c}{Log-Likelihood} \\
& $\log_{10}(A_0)$ & $B_0$ & $\log_{10}(A_0)$ & $B_0$ & $\log_{10}(A_0)$ & $B_0$ \\ \midrule
Number of Parameters, $N_p$ & -19.47 & 0.042 & -22.64 & 0.036 & 4.33 & 0.151 \\
Training Compute, $\mathcal{C}$ & -38.88 & 0.031 &  -43.03 & 0.028 & -6.65 & 0.101 \\
Dataset Size, $\mathcal{D}$ & -8.91 & 0.062 & -30.42 & 0.027 & 7.00 & 0.188 \\ \bottomrule
\end{tabularx}
\end{table}

\section{In-Sequence Predictions to Forecasting}
\label{app:forecast}

Here we simply show an example of how in-sequence test loss correlates with forecasting prediction from roll-out. In particular, in Fig.~\ref{fig:forecast} we show forecasts for three different datasets as a function of model size. Here we use the best weights (i.e., the model that achieved the lowest test loss during training) for each model size and show both in-sequence and forecasting along with the true data. For both the in-seqeunce and forecasting predictions we show the $1\sigma$ range of predictions. Although not perfect, it's clear that as one scales up model size (and therefore in-sequence test loss decreases), forecasting performance also improves substantially. Although we only show three examples here, we observe a similar trend in the forecasting power of our models for a variety of datasets. We leave a more detailed exploration to future work.

\end{document}